\documentclass[journal]{IEEEtran}

\usepackage{amsmath,amssymb}
\usepackage{bm}
\usepackage{graphicx}
\usepackage{booktabs}
\usepackage{array}
\newcolumntype{L}[1]{>{\raggedright\arraybackslash}p{#1}}
\usepackage[hidelinks]{hyperref}
\usepackage{cite}
\usepackage{algorithm}
\usepackage{algpseudocode}

\newtheorem{remark}{Remark}

\newenvironment{standfirst}
  {\par\noindent\bfseries\ignorespaces}
  {\par\vspace{0.6\baselineskip}}

\newcommand{\R}{\mathbb{R}}
\newcommand{\norm}[1]{\lVert #1 \rVert}
\newcommand{\vs}{\mathbf{s}} \newcommand{\ve}{\mathbf{e}}
\newcommand{\vx}{\mathbf{x}} \newcommand{\vy}{\mathbf{y}}
\newcommand{\vp}{\mathbf{p}}
\newcommand{\vJ}{\mathbf{J}} \newcommand{\vJh}{\hat{\mathbf{J}}}
\newcommand{\vq}{\mathbf{q}} \newcommand{\vr}{\mathbf{r}}
\newcommand{\vu}{\mathbf{u}}
\newcommand{\dq}{\Delta\vq}
\newcommand{\vd}{\mathbf{d}}
\newcommand{\vphi}{\bm{\phi}} \newcommand{\vpsi}{\bm{\psi}}
\newcommand{\vW}{\mathbf{W}}
\newcommand{\ds}{\Delta\vs}

\usepackage[most]{tcolorbox}

\newtcolorbox{runningexample}[1]{
    enhanced,
    breakable,
    colback=white,
    colframe=DiagBlue,
    boxrule=0pt,
    leftrule=1.4pt,
    arc=0pt,
    outer arc=0pt,
    left=6pt,
    right=2pt,
    top=4pt,
    bottom=4pt,
    before skip=6pt,
    after skip=6pt,
    before upper={
        \textcolor{DiagBlue}{\bfseries Running example: #1.}\quad
    },
}

\newtcolorbox{intuitions}[1]{
    enhanced,
    colback=white,
    colframe=IntuitionOrange,
    boxrule=0pt,
    leftrule=1.4pt,
    arc=0pt,
    outer arc=0pt,
    left=6pt,
    right=2pt,
    top=4pt,
    bottom=4pt,
    before skip=6pt,
    after skip=6pt,
    before upper={
        \textcolor{IntuitionOrange}{\bfseries Practical Interpretation: #1.}\quad
    },
}

\definecolor{DiagBlue}{RGB}{42,120,214}
\definecolor{GripLeft}{RGB}{74,58,167}
\definecolor{GripRight}{RGB}{235,104,52}
\definecolor{IntuitionOrange}{RGB}{173,92,25}

\graphicspath{{figs/}}

\def\CodeURL{https://github.com/nachocz/tutorial_on_shape_servoing_of_deformable_objects}
\def\DemosURL{https://nachocz.github.io/tutorial_on_shape_servoing_of_deformable_objects/}
\newcommand{\CodeLink}{\expandafter\url\expandafter{\CodeURL}}
\newcommand{\DemosLink}{\expandafter\url\expandafter{\DemosURL}}
\newcommand{\linksragged}{\rightskip=0pt plus 3em\relax}
\usepackage[protrusion=false,expansion=true]{microtype}
\AtBeginDocument{\microtypesetup{expansion=false}}
\newcommand{\AEIcode}{\mbox{MICIU/}\allowbreak\mbox{AEI/}\allowbreak
  \mbox{10.13039/}\allowbreak\mbox{501100011033}}
\newcommand{\FundingAcknowledgment}{\section*{Acknowledgment}
{\microtypesetup{expansion=true}%
This work was supported by the REMAIN project S1/1.1/E0111 (Interreg Sudoe
Programme, ERDF); by project \mbox{PID2024-159279OB-I00}, funded by \AEIcode{}
and by ERDF/EU; by the Gobierno de Arag\'on, group T77\_26R; by grant
\mbox{RYC2024-051408-I}, funded by \AEIcode{} and by ESF+; by the Horizon
Europe project IntelliMan under grant number 101070136; by the SiMOD
project---Multi-robot Mobile Systems for Deformable Objects Manipulation (PR
FESR 21-27 Action 1.1.2, CUP J47G22000830003); and by the French National
Research Agency (ANR) under the France 2030 program with the reference
\mbox{ANR-25-PERO-0003} (PEPR Robotique -- PC DRMI). A.~Caporali is funded by
FSE+ (CUP J19J23000730002).\par}}

\begin{document}

\title{A Tutorial on Shape Servoing of Deformable Objects---Part~II: Advanced
Concepts}

\author{Ignacio~Cuiral-Zueco, Alessio~Caporali, Miguel~Aranda,\\
Gianluca~Palli, Gonzalo~L\'opez-Nicol\'as, and Youcef~Mezouar%
\thanks{This work has been submitted to the IEEE for possible publication.
Copyright may be transferred without notice, after which this version may no
longer be accessible.}%
\thanks{I.~Cuiral-Zueco, M.~Aranda, and G.~L\'opez-Nicol\'as are with the
Instituto de Investigaci\'on en Ingenier\'ia de Arag\'on (I3A), University of
Zaragoza, Spain. A.~Caporali and G.~Palli are with DEI, University of Bologna,
Italy. Y.~Mezouar is with Institut Pascal, Clermont Auvergne INP, France.
Corresponding author: I.~Cuiral-Zueco (e-mail: ignaciocuiral@unizar.es).}}
\markboth{Preprint version}%
{A Tutorial on Shape Servoing of Deformable Objects---Part~II}

\maketitle

\begin{standfirst}
In robotics, \textit{shape control}, or \textit{shape servoing}, refers to the
automatic manipulation of a deformable object toward a desired shape. This
article is the second of a two-part tutorial on data-driven, Jacobian-based
shape control, in which the deformation Jacobian, a local model of how robot
motion changes the measured shape, is estimated from measurements. Part~I built
the shape-servoing loop and derived a local stability condition. This paper
starts by using the singular value decomposition of the deformation Jacobian to
identify shape changes that can be produced by small robot motions, those that
require large robot motions, and those that are locally unavailable. It then
analyses how the choice of shape features determines how the controller
regulates the object's shape and, therefore, how much the
object deforms on its way to the target. It then describes three ways in which
the shape control loop can fail: loss of controllability, loss of observability, and
occurrence of mechanical instabilities. It closes with shape-control extensions
to time-varying targets, secondary tasks, and three-dimensional objects.
\end{standfirst}


\suppressfloats[t]
\begin{figure}[t]
\centering
    \includegraphics[width=\linewidth]{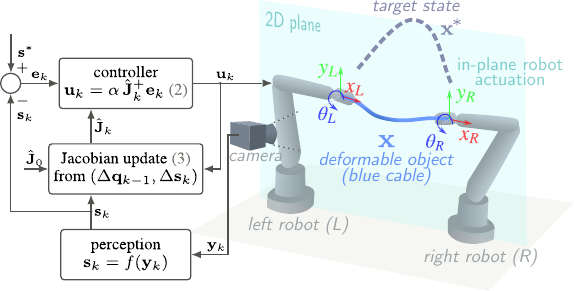}
\caption{Shape-servoing loop of Part~I applied to the running example. Left: the
controller sends the command $\vu_k$ of \eqref{eq:ctrl} to the robots,
perception extracts the features $\vs_k$ from the camera measurement $\vy_k$,
and the Jacobian update corrects the estimate with \eqref{eq:broyden} from the
latest action--response pair, starting from the offline estimate $\vJh_0$.
Right: two robot arms deform an elastic rod $\vx$ (blue), observed by a
stationary camera, toward a target shape $\vx^{*}$ (dashed). The arrows at the
grippers indicate the six in-plane actuation DoFs
$\vq=[{\color{red}x_L},{\color{green}y_L},{\color{blue}\theta_L},
{\color{red}x_R},{\color{green}y_R},{\color{blue}\theta_R}]^{\top}$.}
\label{fig:loopmin}
\end{figure}
\section{Introduction}
\label{sec:intro}

\IEEEPARstart{P}{art~I} of this tutorial developed the complete shape-servoing
loop, from shape feedback and deformation Jacobian estimation to
pseudoinverse-based control. It also established local convergence conditions
and discussed target feasibility and reachability.

Part~II develops the analysis further. We use the singular value decomposition (SVD) of the
deformation Jacobian to group the robot actuation degrees of
freedom (DoFs) into coordinated motion directions and to identify the
independent deformation pattern associated with each one. We
then examine how the choice of shape features determines which deformations can
be observed and regulated, and how losses of controllability, observability, or
mechanical stability can cause shape servoing to fail. Finally, we use these
results to extend the basic control formulation to time-varying targets, secondary control
tasks, and three-dimensional (3-D) objects.

Section~\ref{sec:recap} briefly summarizes the elements of Part~I needed here,
and Table~\ref{tab:roadmap} outlines the contents of Part~II.

\begin{table}[!t]
\caption{Contents of Part~II}
\label{tab:roadmap}
\centering
\renewcommand{\arraystretch}{1.0}
\footnotesize
\setlength{\tabcolsep}{4pt}
\begin{tabular}{@{}L{1.55cm}L{6.15cm}@{}}
\toprule
\textbf{Section} & \textbf{Main content} \\
\midrule
Section~\ref{sec:conditioning} & \emph{Singular value analysis of the
deformation Jacobian}: robot-motion and feature directions, manipulability, and
conditioning. \\
Section~\ref{sec:features} & \emph{Choice of the shape features}: point
coordinates, edge vectors, and curvature-based features; what they measure and
regulate, and how strongly they deform the object. \\
Section~\ref{sec:failures} & \emph{Failure modes of the shape-servoing loop}:
loss of controllability, loss of observability, and mechanical instability;
detection and mitigation. \\
Section~\ref{sec:beyond} & \emph{Extensions of the shape-servoing loop}:
time-varying target shapes, secondary tasks in the null space of the Jacobian
estimate, and 3-D objects. \\
\bottomrule
\end{tabular}
\end{table}

The \textbf{accompanying supplementary material} provides \textbf{public code}
and \textbf{interactive demos},\footnote{\linksragged Code: \CodeLink.
Interactive demos: \DemosLink.} which implement the running example
(Fig.~\ref{fig:loopmin}) and reproduce the numerical results shown in this
article.

\subsection{Summary of Part~I}
\label{sec:recap}

We keep the notation and assumptions of Part~I: manipulation is quasistatic and
controlled at the velocity level, the deformation relevant to the task can be
measured during manipulation, and the relation between robot motion and object
deformation changes smoothly during the task. Figure~\ref{fig:loopmin} shows
the resulting loop. In its perception block, a feature map
$\vs=f(\vy)\in\R^{n}$ extracts the shape representation used for control from
the sensor measurements $\vy$, and the controller reduces the shape error
$\ve=\vs^{*}-\vs$ with respect to a target $\vs^{*}$. The controller outputs a
velocity command $\vu\in\R^{m}$ for the actuated robot coordinates
$\vq\in\R^{m}$, so that $\dot{\vq}=\vu$, and the deformation Jacobian
$\vJ(\vq)=\partial\vs/\partial\vq\in\R^{n\times m}$ gives the local
input--output relation
\begin{equation}
    \dot{\vs}=\vJ(\vq)\,\vu.
    \label{eq:fwd}
\end{equation}

The controller uses an estimate $\vJh\approx\vJ$ and the control law
\begin{equation}
    \vu=\alpha\,\vJh^{+}\ve,
    \label{eq:ctrl}
\end{equation}
where $\alpha>0$ is the control gain and $(\cdot)^{+}$ denotes the
Moore--Penrose pseudoinverse. The loop runs at discrete iterations $k$ with
period $\Delta t$. The latest robot increment $\dq_{k-1}=\vu_{k-1}\Delta t$
and the resulting feature change $\ds_k=\vs_k-\vs_{k-1}$ form an
\emph{action--response pair}, from which the Jacobian update block of
Fig.~\ref{fig:loopmin} corrects the estimate with Broyden's rule
\cite{broyden1965}, using the prediction error
$\vr_k=\ds_k-\vJh_{k-1}\dq_{k-1}$:
\begin{equation}
    \vJh_k
    =
    \vJh_{k-1}
    +
    \frac{\vr_k\,\dq_{k-1}^{\top}}
         {\dq_{k-1}^{\top}\dq_{k-1}}.
    \label{eq:broyden}
\end{equation}
Practical implementations skip this update when $\norm{\dq_{k-1}}$ falls below
a minimum threshold, where $\norm{\cdot}$ denotes the Euclidean norm, and
reject corrections that are implausibly large. The initial estimate $\vJh_0$
is obtained by least-squares regression over small exploratory robot motions
or from a model of the object.

Part~I also established two results that this article uses. First, for a fixed
target, the shape error decreases locally whenever
$\ve^{\top}\vJ\vJh^{+}\ve>0$, i.e., when the feature change produced by the
command has a component along the error, so the estimate must preserve the
corrective directions of the true Jacobian rather than reproduce it entry by
entry. When $n\leq m$ and $\vJ$ has full row rank, the complete shape error
converges locally; when $n>m$, only its component in
$\operatorname{Range}(\vJ)$, the set of locally producible feature velocities,
is corrected. Second, a target is \emph{feasible} if it belongs to the set
$\mathcal{S}\subseteq\R^{n}$ of feature vectors that the robot--object system
can realize, and \emph{reachable} if an admissible manipulation path leads to
it from the current configuration. The Jacobian characterizes only local
reachability: the robot can vary at most
$\operatorname{rank}(\vJ)\leq\min(n,m)$ independent feature directions
locally.

\begin{figure*}[t]
    \centering
    \includegraphics[width=0.92\textwidth]{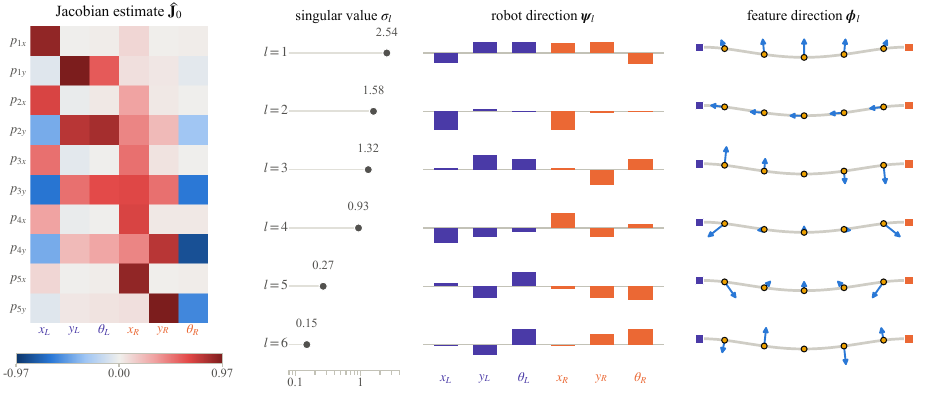}
    \caption{Offline Jacobian estimate of the running example and its SVD. Left:
    signed entries of $\vJh_0$; columns correspond to the coordinates of the
    left (violet) and right (orange) grippers, rows to the image coordinates of
    the tracked points. Right: one row per singular triplet, from the largest to
    the smallest singular value. Each row shows the singular value
    $\sigma_l$ (the directional input--output gain) on a logarithmic axis, the
    robot direction $\vpsi_l$ over the six actuation directions, and the feature
    direction $\vphi_l$, drawn on the rod as one arrow per tracked point
    (yellow). Their relationship is $\vJh_0\vpsi_l=\sigma_l\vphi_l$, where the
    components of $\vpsi_l$ weight the columns of the Jacobian estimate, and
    $\vphi_l$ gives the resulting feature direction with magnitude scaled by
    $\sigma_l$. All rows use the same arrow scale.}
    \label{fig:jacobian}
\end{figure*}

\begin{runningexample}{Setup}
Two robot arms move their grippers in a plane to deform an elastic rod
observed by a stationary camera (Fig.~\ref{fig:loopmin}, right). Each gripper
has three DoFs, two translations and one rotation, so
$\vq=[{\color{red}x_L},{\color{green}y_L},{\color{blue}\theta_L},
{\color{red}x_R},{\color{green}y_R},{\color{blue}\theta_R}]^{\top}$ and
$m=6$. The rod is modeled as a Kirchhoff rod with stretching and bending
stiffness, discretized into $25$ nodes and subject to gravity; its
quasistatic equilibrium minimizes the total potential energy for each gripper
configuration. Perception tracks $P=5$ material points
$\vp_i=[x_i,y_i]^{\top}$ along the rod (yellow dots in
Figs.~\ref{fig:jacobian} and~\ref{fig:features}) and stacks their image
coordinates, so $n=10$. The initial estimate $\vJh_0\in\R^{10\times6}$ is
obtained by regression over $C=24$ exploratory motions, four per DoF.
\end{runningexample}

\section{Singular Value Analysis of the Deformation Jacobian}
\label{sec:conditioning}

In the controller block of Fig.~\ref{fig:loopmin}, the robot command is obtained
by multiplying the shape error by $\vJh^{+}$. Part~I introduced the SVD of
$\vJh$ and its condition number. Here, we further analyze singular
directions to identify the shape changes that, at a specific configuration, can
be produced by small robot motions, those that require large robot motions, and those
that are locally unavailable.

\subsection{Singular Directions and Gains}
\label{sec:svd}

Let
\begin{equation}
    \vJh
    =
    \bm{\Phi}\bm{\Sigma}\bm{\Psi}^{\top}
    =
    \textstyle\sum_{l}\sigma_l\,\vphi_l\vpsi_l^{\top}
    \label{eq:svd}
\end{equation}
be the SVD of the estimate, where the orthogonal matrices
$\bm{\Phi}\in\R^{n\times n}$ and $\bm{\Psi}\in\R^{m\times m}$ have columns
$\vphi_l$ and $\vpsi_l$, and $\bm{\Sigma}\in\R^{n\times m}$ holds the singular
values $\sigma_1\geq\sigma_2\geq\dots\geq0$ on its diagonal. Each right singular
vector $\vpsi_l$ combines several robot DoFs into a robot-motion direction, the
corresponding left singular vector $\vphi_l$ gives the feature direction that
such motion produces, and $\sigma_l$ is the directional
input--output gain between them: a unit robot
velocity along $\vpsi_l$ produces the feature velocity $\sigma_l\vphi_l$
(Fig.~\ref{fig:jacobian}, right). The number of nonzero singular values, i.e.,
the rank of $\vJh$, is the number of independent feature directions that the
robot can vary locally.

The singular vectors also provide orthonormal bases for the subspaces used in
Part~I. The feature directions $\vphi_l$ with $\sigma_l>0$ span
$\operatorname{Range}(\vJh)$, which contains the feature velocities that robot
motion can produce locally according to the estimate $\vJh$; the remaining
feature directions span $\operatorname{Null}(\vJh^{\top})$, which contains the
error components that produce no robot commands. The robot directions $\vpsi_l$
associated with zero singular values span $\operatorname{Null}(\vJh)$. According
to the estimate, motions in this subspace leave the shape features unchanged.
When $n<m$, at least $m-n$ robot directions lie in this null space; these
\emph{free} actuation directions can be used for secondary tasks
(Section~\ref{sec:ext_secondary}).

The pseudoinverse inverts each nonzero gain separately,
\begin{equation}
    \vJh^{+}
    =
    \sum_{\sigma_l>0}\frac{1}{\sigma_l}\,\vpsi_l\vphi_l^{\top},
    \label{eq:pinv}
\end{equation}
so the control law \eqref{eq:ctrl} decomposes the shape error along the feature
directions $\vphi_l$: for each component $\vphi_l^{\top}\ve$, it commands the
robot velocity $\alpha\,\vphi_l^{\top}\ve/\sigma_l$ along the matching robot
direction $\vpsi_l$. An error along a strong feature direction (large
$\sigma_l$) therefore produces a small robot command, an error along a weak
feature direction (small $\sigma_l$) produces a large one, and an error in
$\operatorname{Null}(\vJh^{\top})$ produces none.

\begin{intuitions}{Singular Directions}
The SVD analysis groups the individual robot actuation DoFs into coordinated
motion directions $\vpsi_l$. Each direction $\vpsi_l$ produces a corresponding
independent direction of shape change $\vphi_l$, in the sense that each
$\vphi_l$ represents a shape change pattern that cannot be reproduced by
combining the other directions $\vphi_j$, $j\neq l$. The associated singular
value $\sigma_l$ measures the amount of shape change along $\vphi_l$ produced
per unit of robot motion along $\vpsi_l$. Large $\sigma_l$ indicate a strong
shape change response, while small $\sigma_l$ mean that larger robot motions are
needed to produce the same amount of shape change.
\end{intuitions}

\begin{figure*}[t]
    \centering
    \includegraphics[width=\textwidth]{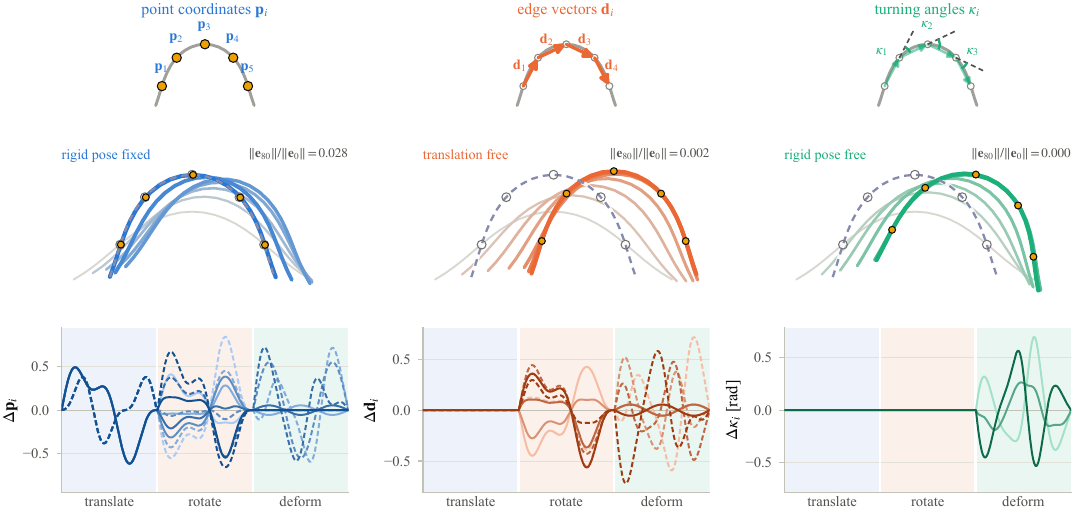}
    \caption{Effect of the feature representation, with one column per
    representation. Top: point coordinates $\vp_i$, edge vectors $\vd_i$, and
    turning angles $\kappa_i$ computed from the same tracked points. Middle:
    closed-loop runs from a common initial configuration (light) to the final
    configuration (dark), with a disturbance that pulls the grippers during the
    first $20$ of $80$ iterations; the dashed curve with grey rings is the
    common target (in the left column, it coincides with the final rod, which
    reaches it), and the label at the top right gives the final normalized error
    $\norm{\ve_{80}}/\norm{\ve_0}$. Bottom: feature components under rigid
    translation, rigid rotation, and non-rigid deformation. Lines are shaded
    from light ($i=1$) to dark along the rod; for $\vp_i$ and $\vd_i$, solid and
    dashed lines denote the $x$ and $y$ components, respectively; the turning
    angles are scalars.}
    \label{fig:features}
\end{figure*}

\subsection{Manipulability, Conditioning, and Scaling}
\label{sec:cond}

Two scalar quantities obtained from the singular values, the manipulability $w$
and the condition number,
\begin{equation}
    w=\textstyle\prod_l \sigma_l,
    \qquad
    \operatorname{cond}
    =
    \sigma_{\max}/\sigma_{\min},
    \label{eq:cond}
\end{equation}
quantify the overall magnitude and the directional anisotropy of the feature
response to robot motion, respectively \cite{yoshikawa1985}, where
$\sigma_{\max}=\sigma_1$ and $\sigma_{\min}$ are the largest and smallest
singular values. Considering \eqref{eq:pinv}, the condition number can also be
interpreted as the ratio between the \textit{largest} and \textit{smallest}
robot velocities induced by a unit-norm feature error aligned with the weakest
and strongest feature directions, respectively.

\begin{remark}[Scaling and uniqueness]
\label{rem:scaling}
Both $w$ and $\operatorname{cond}$ depend on the coordinate units.
For example, changing angular coordinates from radians to degrees rescales the
corresponding columns of $\vJh$ and changes its singular values. Comparisons
therefore require consistent units or a normalization such as
$\vW_{\vs}\vJh\vW_{\vq}^{-1}$, with diagonal scaling matrices $\vW_{\vs}$ and
$\vW_{\vq}$. The sign of a singular vector carries no information, as each pair
$(\vphi_l,\vpsi_l)$ can be multiplied by $-1$. When two singular values are
equal, their individual singular vectors are not unique, but the subspace they
span is uniquely defined. If the singular values are equal or nearly equal,
small perturbations can substantially change the individual vectors while
leaving their joint subspace nearly unchanged.
\end{remark}

\begin{runningexample}{Jacobian Directions}
The heat map of Fig.~\ref{fig:jacobian} (left) shows that each gripper
coordinate mainly moves the tracked points on its own side of the rod, with an
influence that decays toward the opposite end. The vertical translation
${\color{GripLeft}y_L}$, for example, moves $\vp_1$ upward almost one to one
(entry $0.97$) but hardly moves $\vp_5$ (entry $0.03$), and the wrist rotation
${\color{GripLeft}\theta_L}$ acts most strongly on $\vp_2$ and $\vp_3$. Because
the initial configuration is symmetric, the columns of the right-gripper
coordinates mirror those of the left-gripper coordinates. The estimate has full
column rank and $\operatorname{cond}\approx17$, so $\operatorname{Null}(\vJh_0)$
is trivial, which means that, according to $\vJh_0$, every robot motion produces
changes in the shape features. In the SVD analysis (Fig.~\ref{fig:jacobian},
right), the two strongest feature directions, with $\sigma_1=2.54$ and
$\sigma_2=1.58$, move all five points in the same direction, upward and
sideways, respectively, and the third rotates the rod about its center. The
weakest feature direction, with $\sigma_6=0.15$, bends the rod into an S-shape
(arrows alternating down and up along the rod), and the corresponding robot
direction $\vpsi_6$ consists mainly of the wrist rotations (tall bars at
${\color{GripLeft}\theta_L}$ and ${\color{GripRight}\theta_R}$). By
\eqref{eq:pinv}, an error along this bending direction produces a robot command
about $17$ times larger than an equal error along the strongest feature
direction.
\end{runningexample}

The SVD analysis shows which feature directions are weak or unavailable, but not
why. The Jacobian $\vJ$ couples two effects: how robot motion deforms the object
and how the selected features register that deformation. A small or zero
singular value can therefore come from robot motion that produces little or no
deformation, or from deformation that produces little or no change in the
features. Sections~\ref{sec:features} and~\ref{sec:failures} examine these two
effects, together with mechanical instabilities, and their consequences for the
closed loop.

\section{Choosing the Shape Features}
\label{sec:features}

To keep Part~I simple and self-contained, the feature map $\vs_k=f(\vy_k)$ was
restricted to the image coordinates of tracked material points. Here, we
consider alternative shape representations that differ in geometric structure
and examine how this choice determines 1) which shape changes are
\textit{perceivable} by the control loop and 2) the \textit{deformation paths}
that the object follows during shape servoing.

In particular, we compare three representations built from the same $P$ tracked
points $\vp_1,\dots,\vp_P$ (Fig.~\ref{fig:features}, top row):

\begin{itemize}
    \item \textbf{Point coordinates:}
    $\vs=[\vp_1^\top,\dots,\vp_P^\top]^\top\in\R^{2P}$. They retain both the
    shape and the pose of the object and are therefore \emph{extrinsic} features
    in the terminology of \cite{cuiral2024}. They extend directly to 3-D points.

    \item \textbf{Edge vectors:} $\vd_i=\vp_{i+1}-\vp_i$, $i=1,\dots,P-1$. They
    remove absolute position but retain the relative displacements and the
    global orientation of the object, so they are invariant to translation but
    not to rotation. In 3-D, analogous quantities are defined on the edges of
    surface or volumetric meshes.

    \item \textbf{Curvature-based features:} the turning angles $\kappa_i$,
    $i=1,\dots,P-2$, are the signed angles between consecutive edges $\vd_i$ and
    $\vd_{i+1}$, wrapped to $(-\pi,\pi]$. They describe local bending, are
    invariant to rigid translation and rotation, and are therefore
    \emph{intrinsic} features. 3-D counterparts include dihedral angles,
    discrete curvature vectors, and Laplacian or mean-curvature coordinates on
    surface meshes.
\end{itemize}

Table~\ref{tab:invariance} and the bottom row of Fig.~\ref{fig:features}
illustrate the invariances of the three representations: a rigid translation
affects only point coordinates, a rigid rotation leaves only curvature-based
features unchanged, and a non-rigid deformation affects all three
representations.

\begin{intuitions}{Feature Choice}
The shape representation determines which changes of the object alter the
feature vector, and therefore appear in the error, and which leave it unchanged.
These invariances are useful when they \textit{remove} motions that are
irrelevant to the task \cite{cuiral2024}, but harmful when they \textit{hide}
quantities that we want to regulate.
\end{intuitions}

\begin{table}[t]
\caption{Feature Maps, Their Invariances, and What the Controller Can Regulate}
\label{tab:invariance}
\centering
\renewcommand{\arraystretch}{1.2}
\scriptsize
\setlength{\tabcolsep}{3.2pt}
\begin{tabular*}{\linewidth}{@{\extracolsep{\fill}}lccccl@{}}
\toprule
\textbf{Feature map} & $n$ & \textbf{Transl.} & \textbf{Rot.} & \textbf{Type} &
\textbf{Free} \\
\midrule
Point coords.\ $\vp_i$    & $2P$     & ---  & ---  & extrinsic & nothing \\
Edge vectors $\vd_i$      & $2(P-1)$ & inv. & ---  & mixed     & translation \\
Curvature $\kappa_i$      & $P-2$    & inv. & inv. & intrinsic & rigid motion \\
\bottomrule
\end{tabular*}
\\[2pt]
\begin{minipage}{\linewidth}
\raggedright\scriptsize ``inv.'' denotes invariance to the indicated rigid
transformation. Matching the target feature vector therefore does not constrain
the transformation listed in the ``Free'' column. Dimensions are for $P$ tracked
points in the plane.
\end{minipage}
\end{table}

\begin{runningexample}{Feature Invariances}
In the middle row of Fig.~\ref{fig:features}, a temporary disturbance,
representative of those that may occur in a real setup, pulls the grippers and
tends to move the rod rigidly away from the target pose. Point coordinates
register this motion as an error, so the loop resists the disturbance and
returns the rod to the target pose once it stops. Edge vectors are invariant to
translations, while curvature-based features are also invariant to rotations.
The loop therefore does not correct these components of the rigid-body motion,
and the rod reaches the target shape translated (for edge vectors), or
translated and rotated (for curvature-based features), relative to the target
pose.
\end{runningexample}

The same invariances appear in the deformation Jacobian: any robot velocity that
moves the object only in ways to which the features are invariant lies in
$\operatorname{Null}(\vJ)$ (Section~\ref{sec:svd}).

\begin{figure}[t]
    \centering
    \includegraphics[width=\columnwidth]{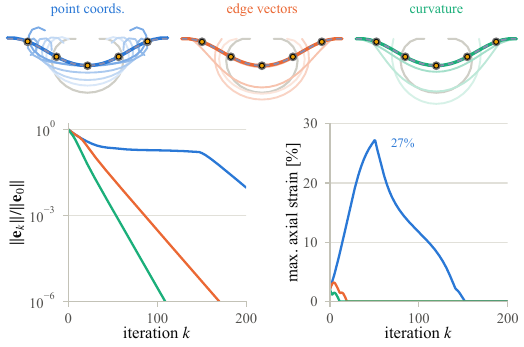}
    \caption{Shape servoing of a rod from a ring-shaped geometry to a less
    curved target shape, with each shape representation ($P=5$; all runs use the
    damped inverse of Section~\ref{sec:cat1} with $\mu=0.05$). Top: rod at $k=0$
    (grey ring), at $k=5$, $10$, $20$, $50$, and $100$ (light to dark), and at
    $k=200$ (darkest); the dashed curve is the target. Bottom: normalized
    feature error and maximum axial strain over the rod edges.}
    \label{fig:unloop}
\end{figure}

Feature choice also affects the path that the object follows toward the target.
Point coordinates represent each material point independently, whereas edge
vectors and curvature-based features also encode the connectivity between
adjacent material points: edge lengths relate to local stretching, and turning
angles to local bending. Reducing the error in these spaces therefore acts more
directly on the material deformation than reducing the point-coordinate error,
which measures image distances between points regardless of how they are
connected through the material. The resulting paths can be more compliant with
the object (Fig.~\ref{fig:unloop}); the $SE(2)$ formulation of
\cite{cuiral2023shapecompliant} builds on this idea.

\begin{runningexample}{Looped Rod}
In Fig.~\ref{fig:unloop} all three shape representations allow the controller to
reach the same target shape. With point coordinates as features, the rod bends
sharply near the grippers, and its maximum axial strain reaches $27\%$. With
curvature-based features, the rod opens first, and its strain never exceeds its
initial value of $2.1\%$; with edge vectors, it peaks at $3\%$. This comparison shows
how strongly feature choice can affect the deformation imposed on the way to the
target. Note that, for curvature-based features, $P=5$ and $m=6$,
so $n=P-2=3<m=6$, and at least three robot directions lie in
$\operatorname{Null}(\vJ)$, among them the gripper motions that translate or
rotate the rod rigidly.
\end{runningexample}

Feature types can also be combined. Richer representations do so at each
material point: local poses in the special Euclidean groups $SE(2)$ or $SE(3)$,
for example, associate each point's position with a local orientation.
Alternatively, different feature vectors can be stacked, with suitable weights
when their units or scales differ, or arranged hierarchically, as in
Section~\ref{sec:ext_secondary}. Other representations in the literature include
points sampled along cables \cite{lagneau2020}, Fourier descriptors for contours
\cite{navarro2018}, and mesh- or lattice-based features for surfaces and solids
\cite{shetab2022,yin2021}. Image moments play an analogous role in visual
servoing \cite{chaumette2007}.

\begin{remark}[Modal shape representations]
Modal representations, such as Fourier coefficients or principal component
analysis (PCA) modes, express the same feature information in a different basis.
They add no geometric or physical structure, but they allow the components to be
ranked and truncated by frequency, variance, or energy, which retains the
dominant shape information and filters out finer detail.
\end{remark}

As introduced in Part~I, the shape representations considered here require a
correspondence between points on the current and target shapes. This
correspondence, or \emph{domain map}, ensures that matching feature entries
refer to the same material points on the object, so that the shape error has a
consistent physical meaning. For tracked material points, for example with
fiducial markers, this correspondence holds by construction. For detected
contours and surfaces without persistent point identities, however, the domain
map must be estimated. Deformation-aware methods construct such maps by
minimizing a deformation cost (e.g., elastic energy) between the mapped current
and target shapes, as in \cite{CUIRALZUECO2025105134} for contours and
\cite{cuiral2025time} for surfaces.

\section{Failure Modes of the Shape-Servoing Loop}
\label{sec:failures}

The deformation Jacobian also helps explain why shape servoing can fail. The
chain-rule factorization introduced in Part~I,
\begin{equation}
    \vJ \;=\; \frac{\partial \vs}{\partial \vx}\,
              \frac{\partial \vx}{\partial \vq},
    \label{eq:chain}
\end{equation}
where $\vx$ denotes the full configuration of the object, separates how robot
motion changes this configuration from how those changes appear in the selected
features. A \emph{controllability failure}, or loss of control authority, occurs
when $\partial\vx/\partial\vq$ loses rank, so that some robot motions can no
longer generate independent object deformations. An \emph{observability
failure}, or loss of shape sensitivity, occurs when an object motion lies in
$\operatorname{Null}(\partial\vs/\partial\vx)$ and therefore produces no change
in the selected features. Controllability and observability thus refer to these
local relations in the shape-servoing loop rather than to their classical
state-space definitions.

The shape-servoing loop estimates the Jacobian $\vJ$ in \eqref{eq:chain}, which
combines the robot-to-object and object-to-feature mappings. Since $\vJ$ does
not uniquely determine its factors, controllability and observability failures
cannot generally be distinguished from $\vJ$ alone. Furthermore, \eqref{eq:fwd}
presupposes that the underlying quasistatic equilibrium relation remains regular.
When this regularity is lost, as at a buckling or snap-through transition, the
resulting failure is a \emph{mechanical instability}. Table~\ref{tab:failures}
summarizes these three failure modes, their observable signatures, and possible
mitigation strategies.

\begin{intuitions}{Failure Diagnosis}
When a robot motion produces little or no feature change, the motion may have
produced little deformation, for example when a rod is pulled taut, or the
object may have moved in a way that the selected features do not capture, such
as a pure translation not being reflected in curvature-based features. The
measured response is similar in both cases, so distinguishing the underlying
cause requires additional information (e.g., a different feature choice or a mechanical
model).
\end{intuitions}

\begin{figure}[t]
    \centering
    \includegraphics[width=\columnwidth]{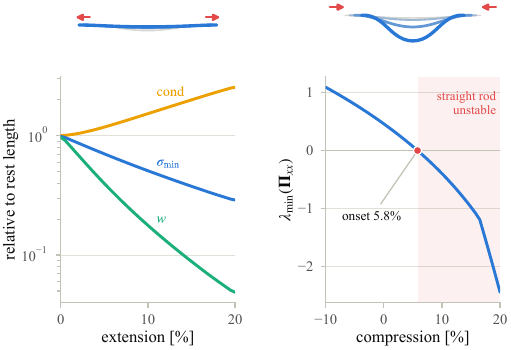}
    \caption{Examples of loss of control authority and mechanical instability.
    Left: as the rod is stretched beyond its rest length, the smallest singular
    value $\sigma_{\min}$ and manipulability $w$ of the exact Jacobian decrease,
    while its condition number increases. Right: during compression, the
    smallest eigenvalue of the potential-energy Hessian of the straight rod
    crosses zero at the onset of buckling (red dot); beyond this point, the
    straight configuration is unstable (shaded). The strips above the plots show
    the corresponding rod configurations and gripper motions.}
    \label{fig:failures}
\end{figure}

\begin{table}[t]
\caption{Failure Modes, Observable Signatures, and Mitigation}
\label{tab:failures}
\centering
\renewcommand{\arraystretch}{1.35}
\footnotesize
\setlength{\tabcolsep}{3pt}
\begin{tabular}{@{}L{1.72cm}L{1.90cm}L{1.72cm}L{1.72cm}@{}}
\toprule
\textbf{Failure} & \textbf{Underlying cause} & \textbf{Observable signature} &
\textbf{Mitigation} \\
\midrule
Controllability (taut, slack) & $\partial\vx/\partial\vq$ loses rank & weak
singular directions; $\dq_{k-1}\neq\mathbf{0}$, $\ds_k\approx\mathbf{0}$ &
damped inverse \cite{nakamura1986}; monitor $\sigma_{\min}$ and $w$ \\

Observability (invariance, occlusion) & motion in
$\operatorname{Null}(\partial\vs/\partial\vx)$ & weak singular directions;
$\dq_{k-1}\neq\mathbf{0}$, $\ds_k\approx\mathbf{0}$ & change feature map or
viewpoint \\

Mechanical instability (buckling, snap-through) & quasistatic equilibrium loses
regularity & small $\norm{\dq_{k-1}}$ with large or poorly predicted
$\norm{\ds_k}$ & fully re-estimate $\vJh$ after poor predictions \\
\bottomrule
\end{tabular}
\end{table}

\subsection{Loss of Controllability}
\label{sec:cat1}

As control authority is lost, some robot motions produce progressively smaller
changes in the object configuration. This can happen when the object is pulled
taut, as in the example below, but also when it is slack: if a rod lies on a
table, for instance, small gripper motions may move only the segment near the
gripper while the far end does not respond. In the deformation Jacobian, one or
more singular values then decrease: at fixed coordinate scaling, $\sigma_{\min}$
approaches zero, $w$ decreases, and the condition number grows. The factor
$1/\sigma_{\min}$ in the pseudoinverse \eqref{eq:pinv} can then generate large
robot commands along directions that produce little deformation. For a physical
system holding a stretched object, such commands can be both ineffective and
unsafe.

\begin{runningexample}{Taut Rod}
The left plot of Fig.~\ref{fig:failures} illustrates this behavior as the
grippers pull the rod taut from the configuration of Fig.~\ref{fig:jacobian}.
Beyond the rest length, gripper translations still move the rod, while wrist
rotations produce progressively less deformation. The weakest robot direction
$\vpsi_{\min}$, a unit vector, becomes increasingly a wrist rotation: the
squares of its two wrist entries sum to $0.68$ at the rest length and to $0.78$
at $20\%$ extension. Over the same interval, $\sigma_{\min}$ decreases from
$0.15$ to $0.04$, $w$ decreases by a factor of $21$, and the condition number
increases from $16$ to $42$. These values are computed from the exact Jacobian
of the simulated rod, so the trend reflects the mechanics rather than estimation
error; at the rest length, the estimate $\vJh_0$ of Section~\ref{sec:cond} gives
a similar $\operatorname{cond}\approx17$.
\end{runningexample}

Online, the same loss of sensitivity appears in measurements with
$\dq_{k-1}\neq\mathbf{0}$ but $\ds_k\approx\mathbf{0}$. The Broyden update
\eqref{eq:broyden} adapts the Jacobian estimate accordingly, but small singular
values make its pseudoinverse increasingly ill-conditioned. A standard
mitigation is to replace $\vJh_k^{+}$ by the damped least-squares inverse
$\vJh_k^{\top}(\vJh_k\vJh_k^{\top}+\mu^2\mathbf{I}_n)^{-1}$, where $\mu>0$ is a
damping factor and $\mathbf{I}_n$ is the $n\times n$ identity matrix
\cite{nakamura1986}. This replaces each gain $1/\sigma_l$ in \eqref{eq:pinv} by
$\sigma_l/(\sigma_l^2+\mu^2)\leq1/(2\mu)$, bounding the command amplification
along weak singular directions.

\subsection{Loss of Observability}
\label{sec:cat2}

Loss of observability occurs when changes in the object are not reflected in the
measured features. This may arise from the feature representation itself, as
with the invariances in Fig.~\ref{fig:features}, or from perception, for example
when part of the object becomes occluded.

The consequences of loss of observability can range from negligible to
mechanically critical. For example, with curvature-based features, an unobserved
rigid translation may be irrelevant to the task. In contrast, unobserved
stretching can increase tension in a taut rod and eventually reduce control
authority, while an unobserved axial twist of an untextured rod can accumulate
torsional strain and eventually trigger pigtailing, a mechanical instability
discussed next. Loss of observability can therefore allow the object to approach
other failure modes.

\subsection{Mechanical Instability}
\label{sec:cat3}

Let $\Pi(\vx;\vq)$ denote the total potential energy of the object for a given
gripper configuration $\vq$. A stable equilibrium $\bar{\vx}(\vq)$ is a local
minimum of $\Pi$ with positive-definite Hessian
$\bm{\Pi}_{\vx\vx}=\partial^2\Pi/\partial\vx^2$. As the grippers progressively
load the object, it follows the equilibrium branch $\bar{\vx}(\vq)$, and its
stiffness along one deformation mode may decrease until the smallest eigenvalue
of the Hessian vanishes,
\begin{equation}
    \lambda_{\min}\!\big(\bm{\Pi}_{\vx\vx}\big) \;\to\; 0.
    \label{eq:bifurcation}
\end{equation} As this limit is approached,
small gripper motions can produce increasingly large deformations, and the
quasistatic sensitivity may become unbounded at the bifurcation. The equilibrium
mapping then loses local regularity: snap-through may produce a discontinuous
change in configuration, while supercritical buckling leads to a new equilibrium
branch \cite{bretl2014}. In either case, the local relation \eqref{eq:fwd} does
not remain valid through the transition.

\begin{runningexample}{Buckling}
The right plot of Fig.~\ref{fig:failures} illustrates this transition as the
grippers compress the rod. The smallest eigenvalue of the Hessian along the
straight configuration decreases with compression and crosses zero at $5.8\%$
compression. Beyond this point, the straight configuration becomes unstable, and
the rod buckles, as the strip above the plot shows.
\end{runningexample}

Such transitions also affect online Jacobian estimation. A small $\dq_{k-1}$ may
produce a disproportionately large $\ds_k$, resulting in a large Broyden
correction and an unreliable inverse. The model update safeguard
summarized in Section~\ref{sec:recap} can reject these samples, after which the
Jacobian can be re-estimated around the new equilibrium. If a model of the
object predicts the instability, planning can instead avoid the corresponding
configurations \cite{bretl2014}.

The loop already computes three quantities that can reveal these failures: the
singular values of $\vJh_k$, the prediction error $\vr_k$, and the response
ratio $\norm{\ds_k}/\norm{\dq_{k-1}}$. Small singular values or a small response
ratio indicate weak sensitivity, whereas a large prediction error or an
unusually large response ratio suggests that the local model no longer
represents the object. These quantities can therefore warn that the loop is
approaching a failure, but identifying its physical cause may require additional
sensing or a model of the object.

\section{Extensions of the Shape-Servoing Loop}
\label{sec:beyond}
The loop of Fig.~\ref{fig:loopmin} can be extended without changing its main
structure. We introduce two extensions: trajectory tracking of time-varying
targets and the inclusion of secondary control tasks through redundant robot
motions. These extensions of the control law in \eqref{eq:ctrl} can be used
separately or together. We also show how the concepts developed throughout
Parts~I and~II generalize to 3-D objects and to settings beyond the 2-D running
example (Section~\ref{sec:ext_objects}).

\subsection{Time-Varying Target Shapes}
\label{sec:ext_target}

Part~I considered a fixed target, $\dot{\vs}^{*}=\mathbf{0}$. We now consider
tracking a sequence of desired shapes defined over time, $\vs^{*}(t)$. In this
case, $\dot{\ve}=\dot{\vs}^{*}-\vJ\vu$, so target motion continuously introduces
error. Feedback alone reacts only after this error appears, producing a tracking
lag that a larger gain $\alpha$ can reduce but not eliminate. The target motion
can instead be anticipated with a feedforward term \cite{chaumette2007}:
\begin{equation}
    \vu_k
    =
    \alpha\,\vJh_k^{+}\ve_k
    +
    {\color{DiagBlue}\vJh_k^{+}\dot{\vs}^{*}_k},
    \label{eq:ctrl_ff}
\end{equation}
where $\dot{\vs}^{*}_k$ is the target velocity at iteration $k$. If
$\dot{\vs}^{*}\in\operatorname{Range}(\vJ)$ and the Jacobian estimate is
accurate, the added term reproduces the target motion and removes its
contribution to the tracking error.

\begin{figure}[t]
    \centering
    \includegraphics[width=\columnwidth]{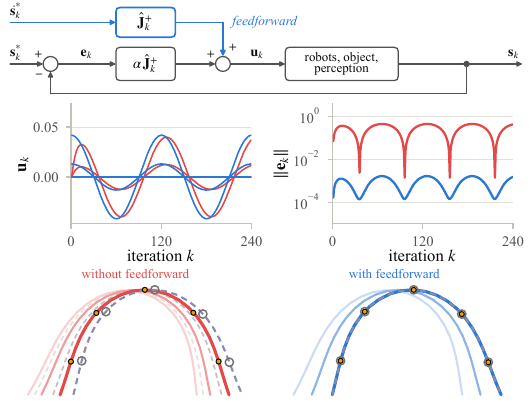}
    \caption{Tracking of a time-varying target in the running example, with and
    without the feedforward term of \eqref{eq:ctrl_ff} ($\alpha=0.15$). Top:
    block diagram of the controller, with the feedforward term in blue; the
    Jacobian update of Fig.~\ref{fig:loopmin} is omitted. Middle, with (blue)
    and without (red) feedforward: components of the robot command $\vu_k$ for
    the left gripper; those of the right gripper coincide by symmetry (left),
    and shape error $\norm{\ve_k}$ in simulation units (right). Bottom: target
    (dashed) and rod without (left) and with (right) feedforward at $k=106$ and
    $116$ (light) and at $k=126$, the instant of largest lag (dark; grey rings:
    target points).}
    \label{fig:feedforward}
\end{figure}

\begin{figure*}[t]
    \begin{minipage}[t]{\columnwidth}\vspace{0pt}\centering
        \includegraphics[width=\columnwidth]{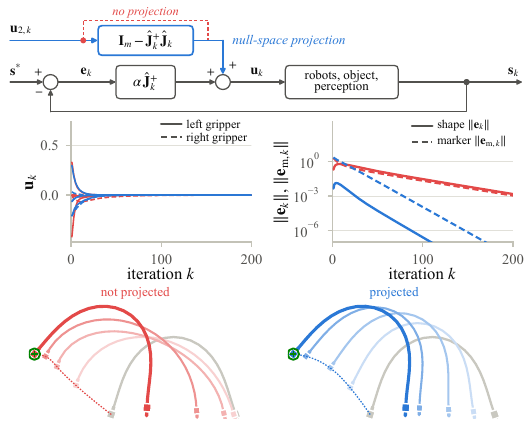}
        \caption{Secondary task in the running example: the left gripper is
        driven toward a marker (green) while curvature-based features ($n=3$)
        hold the arch shape ($\alpha=0.1$, $\beta=0.2$). Top: block diagram of
        \eqref{eq:ctrl_null}, with the projected secondary velocity in blue and
        the unprojected alternative in red; the Jacobian update of
        Fig.~\ref{fig:loopmin} is omitted. Middle, with (blue) and without (red)
        projection: components of $\vu_k$ for the left (solid) and right
        (dashed) grippers (left), and shape error $\norm{\ve_k}$ in radians
        (solid) and marker error $\norm{\ve_{\mathrm{m},k}}$ in simulation units
        (dashed) (right, logarithmic vertical axis). Bottom: rod and grippers
        without (left) and with (right) projection at $k=0$ (grey), at $k=3$,
        $8$, and $20$ (light), and at $k=200$ (dark). The dotted lines near the
        object are the paths followed by the left gripper.}
        \label{fig:nullspace}
    \end{minipage}\hfill
    \begin{minipage}[t]{\columnwidth}\vspace{0pt}\centering
        \includegraphics[width=\columnwidth]{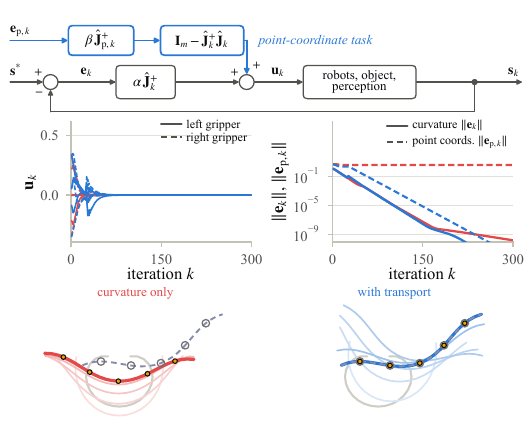}
        \caption{Curvature-based features with a point-coordinate transport
        task, from the initial rod configuration in Fig.~\ref{fig:unloop} to a target
        configuration obtained by translating the target shape in that figure by
        $(1.2,0.6)$ and rotating it by $20^\circ$ ($\alpha=0.12$, $\beta=0.1$, damped
        inverses with $\mu=0.05$ in both tasks). Top: block diagram of
        \eqref{eq:ctrl_null} with
        $\vu_{2,k}=\beta\,\vJh_{\mathrm{p},k}^{+}\ve_{\mathrm{p},k}$ (blue). Middle:
        results with (blue) and without (red) transport. Left: components of $\vu_k$ for
        the left (solid) and right (dashed) grippers. Right: curvature error
        $\norm{\ve_k}$ in radians (solid) and point-coordinate error
        $\norm{\ve_{\mathrm{p},k}}$ in simulation units (dashed), with a logarithmic
        vertical axis. Bottom: rod configurations with curvature-based features alone
        (left) and with transport (right), at $k=0$ (grey), $k=5$, $10$, $15$, $25$,
        $40$, $70$, $100$, and $150$ (light to dark), and $k=300$. The dashed curve with
        grey rings indicates the target configuration.}
        \label{fig:transport}
    \end{minipage}
\end{figure*}

\begin{runningexample}{Moving Target}
In Fig.~\ref{fig:feedforward}, the arch target of Part~I sways horizontally and
tilts with a period of $120$ iterations. Because the target trajectory is known,
we can use $\dot{\vs}^{*}_k$ as the target change over the next control period
divided by $\Delta t$. Without the feedforward term, the shape error oscillates
with peaks of $0.46$, and each tracked point lags about $0.2$ behind its target
position when the target moves fastest. With the feedforward term, the shape
error remains below $1.7\times10^{-3}$.
\end{runningexample}

\subsection{Secondary Tasks in the Null Space of the Deformation Jacobian}
\label{sec:ext_secondary}

When the shape-servoing task leaves some robot motion directions unconstrained,
these directions can simultaneously serve secondary tasks, such as bringing one
gripper to a desired position while the object maintains its shape. These
directions span $\operatorname{Null}(\vJh_k)$, which in that case is nontrivial
(Section~\ref{sec:svd}). If $\vu_{2,k}\in\R^m$ denotes the robot velocity
requested by the secondary task, its projection onto this null space can be
added to the control law:
\begin{equation}
    \vu_k
    =
    \alpha\,\vJh_k^{+}\ve_k
    +
    {\color{DiagBlue}
    \big(\mathbf{I}_m-\vJh_k^{+}\vJh_k\big)\vu_{2,k}},
    \label{eq:ctrl_null}
\end{equation}
where $\mathbf{I}_m$ is the $m\times m$ identity matrix and
$\mathbf{I}_m-\vJh_k^{+}\vJh_k$ projects the secondary command onto
$\operatorname{Null}(\vJh_k)$. Since
$\vJh_k(\mathbf{I}_m-\vJh_k^{+}\vJh_k)=\mathbf{0}$, these additional robot
motions leave the shape features unchanged according to the current Jacobian
estimate. Any residual effect on the shape-servoing task comes from estimation
error or higher-order effects and is corrected by the first term of
\eqref{eq:ctrl_null}.

The number of DoFs available for secondary tasks,
$m-\operatorname{rank}(\vJh_k)$, therefore depends on the number of actuation
DoFs, on the number of features and their invariances, and on the current
configuration. In the running example, point coordinates give a trivial null
space (Section~\ref{sec:cond}). Curvature-based features imply a $3\times6$
Jacobian. Since $\operatorname{rank}(\vJh_k)\leq3$, rank--nullity gives
$\dim\operatorname{Null}(\vJh_k)\geq6-3=3$: at least three independent
combinations of robot DoFs remain available for the secondary task
(Section~\ref{sec:features}).

\begin{runningexample}{Secondary Task}
In Fig.~\ref{fig:nullspace}, curvature-based features hold the arch shape while
the secondary task drives the left gripper toward a marker at
$(m_x,m_y)=(-3.8,2.8)$. Its command acts only on the translation of the left
gripper, $\vu_{2,k}=\beta\,[\ve_{\mathrm{m},k}^{\top},0,0,0,0]^{\top}$, where
$\ve_{\mathrm{m},k}=[m_x-x_{L,k},\,m_y-y_{L,k}]^{\top}$ is the marker error of
the left gripper and $\beta=0.2$ is the gain. With the projection, both grippers
move so that the rod is translated and rotated with an almost unchanged shape:
the shape error remains below $1.4\times10^{-2}$\,rad while the left gripper
reaches the marker. Without the projection, the same secondary velocity deforms
the rod, and the shape error rises to $0.65$\,rad before the first term of
\eqref{eq:ctrl_null} reduces it again.
\end{runningexample}

The null space also allows different feature representations to be combined
hierarchically. For example, curvature-based features can guide the object
toward the target shape along a low-deformation path (Fig.~\ref{fig:unloop}),
while point coordinates additionally constrain its pose
(Fig.~\ref{fig:features}, left column). We therefore use curvature-based
features for the primary task in \eqref{eq:ctrl_null} and define the secondary
point-coordinate command as
$\vu_{2,k}=\beta\,\vJh_{\mathrm{p},k}^{+}\ve_{\mathrm{p},k}$, where $\beta>0$ is
its gain, and $\ve_{\mathrm{p},k}$ and $\vJh_{\mathrm{p},k}$ are the error and
Jacobian estimate of the point coordinates. The null-space projection retains
only the component of $\vu_{2,k}$ that leaves the curvature-based features
unchanged to first order. The curvature-based features therefore take priority
and regulate the shape with little deformation, while the point-coordinate
command uses the remaining freedom of robot motion to regulate the pose. Only
$\vJh_{\mathrm{p},k}$ needs to be estimated, with \eqref{eq:broyden}: the
turning angles $\bm{\kappa}$ are a known function of the point coordinates
$\vs_{\mathrm{p}}$, so the chain rule gives
$\vJh_k=(\partial\bm{\kappa}/\partial\vs_{\mathrm{p}})\,\vJh_{\mathrm{p},k}$.
Robot motions that $\vJh_{\mathrm{p},k}$ predicts to move the tracked points
rigidly therefore lie in $\operatorname{Null}(\vJh_k)$.

In Fig.~\ref{fig:transport}, the rod must transition from its initial
configuration to the target shape and pose. Using curvature-based features alone
recovers the target shape but leaves a point-coordinate error of $3.6$. Adding a
null-space-projected point-coordinate task reduces the point-coordinate and
curvature errors below $10^{-4}$ and $10^{-8}$, respectively, within $300$
iterations, without increasing the maximum axial strain beyond its initial
value.

\subsection{Three-Dimensional Deformable Objects}
\label{sec:ext_objects}

The structure of the loop in Fig.~\ref{fig:loopmin} also applies to 3-D
deformable objects, although some of its blocks require additional
considerations. In particular, perception must provide a feature representation
suitable for the deformation task and maintain correspondence between current
and target features. These features do not need to represent the full 3-D
geometry. A 3-D object can also be controlled through features defined in the
2-D image space, at the cost of leaving some motions unconstrained and therefore
admitting redundant solutions \cite{cuiral2024}. Alternatively, 3-D measurements
can provide a point cloud from which features are extracted, such as points
tracked along a cable \cite{caporali2026robotic} or features defined on a
time-consistent mapping of the object surface \cite{cuiral2025time}.

\begin{figure}[t]
    \centering
    \includegraphics[width=\columnwidth]{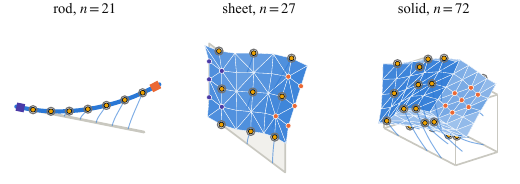}
    \caption{Shape control of three simulated 3-D objects with the control law
    \eqref{eq:ctrl} and the Jacobian of the deformation model. Each object is
    shown at $k=0$ (grey) and at $k=40$ (blue), with the fixed grasp (violet),
    the actuated grasp with $m=3$ coordinates (orange), the feature vertices
    (yellow), their target positions (grey rings), and their paths (thin blue
    lines). The rod is a cubic centerline between two grasped segments; the
    sheet (triangular surface mesh) and the solid (volumetric lattice) deform
    according to an ARAP model. Feature vectors stack the 3-D positions of the
    marked vertices, so $n=21$, $27$, and $72$.}
    \label{fig:spatial}
\end{figure}

Figure~\ref{fig:spatial} illustrates 3-D shape control with a rod, whose
centerline is a cubic curve between its two grasped segments, and with a sheet
and a volumetric object that deform according to an as-rigid-as-possible (ARAP)
model, as used for shape servoing in \cite{shetab2022}. The controller applies
\eqref{eq:ctrl} with $\alpha=0.22$ and a bounded robot step, using the Jacobian
of the simulated model, evaluated by finite differences, instead of an estimate.

\section{Conclusion}
\label{sec:conclusion}

Part~I of this tutorial built a shape-servoing loop that needs no explicit
mechanical model: it estimates the local relation between robot motion and
measured shape change and uses this relation for feedback control. In this
Part~II, we analyzed the structure and the limitations of that loop. We saw that
the SVD of the Jacobian decomposes robot actuation into coordinated motion
directions, each associated with an independent shape-change direction and with
a gain that quantifies how strongly the shape responds. Along directions with
small gains, the pseudoinverse turns even small shape errors into large robot
commands.

We then showed that the feature representation determines both what the
controller regulates and the deformation path that the object follows toward the
target. Point coordinates constrain shape and pose, whereas edge vectors and
curvature-based features introduce invariances and encode quantities related to
stretching and bending. In our looped-rod example, the controller using
curvature-based features opened the loop with little stretching, whereas the
point-coordinate controller produced local strains of up to $27\%$. Through
null-space motion, a hierarchical controller can combine both: one
representation drives the primary shape-servoing task, and the other regulates
the remaining pose freedom.

The same Jacobian formulation also helped us identify how the loop can fail.
Control authority is lost when robot motion produces little deformation,
observability is lost when changes of the object produce little change in the
measured features, and a mechanical instability can invalidate the local
quasistatic relation. The first two failures can produce similar measured
responses, so telling them apart may require additional information, whereas a
mechanical instability can appear as large or poorly predicted shape changes.
The singular values, prediction errors, and response ratios that the loop
already computes can warn of all three.

Finally, we showed how the same loop extends to time-varying targets through a
feedforward term, to secondary tasks through null-space motion, and to 3-D
objects.

Several challenges remain open. Plastic and other history-dependent materials
require models that adapt as the relation between $\vq$ and $\vs$ changes with
the material state and deformation history
\cite{cherubini2020,laezza2021learning}, while discontinuous events such as
object breakage or joining, environmental contacts, and regrasping can abruptly
change the local deformation map and available control authority. A more
fundamental departure occurs when the object no longer responds quasistatically:
shape regulation then remains a servoing problem, but the Jacobian model considered
here is no longer sufficient, and dynamic shape servoing must account explicitly
for effects such as inertia, damping, deformation rates, and internal material
states.

\looseness=-1
We hope that the two parts of this tutorial, together with the accompanying code
and interactive demos, provide a practical starting point for understanding,
implementing, and extending the Jacobian-based shape-servoing framework
developed here.

\FundingAcknowledgment

\IEEEtriggeratref{9}
\bibliographystyle{IEEEtran}
\bibliography{references}

\end{document}